\documentclass{article}

\PassOptionsToPackage{numbers, sort&compress}{natbib}
 \usepackage[eandd, nonanonymous, preprint]{neurips_2026}
\usepackage[utf8]{inputenc} % allow utf-8 input
\usepackage[T1]{fontenc}    % use 8-bit T1 fonts
\usepackage{hyperref}       % hyperlinks
\usepackage{url}            % simple URL typesetting
\usepackage{booktabs}       % professional-quality tables
\usepackage{amsfonts}       % blackboard math symbols
\usepackage{nicefrac}       % compact symbols for 1/2, etc.
\usepackage{microtype}      % microtypography
\usepackage{xcolor}         % colors

\usepackage{multirow}

\usepackage{amsmath, amssymb, amsfonts} % For math symbols
\usepackage{graphicx} % Required for \includegraphics (though not used in this text-only example)
\usepackage{bm} % For bold math symbols using \bm
\usepackage{bbm}
\usepackage{xspace}

\usepackage{booktabs}
\usepackage{array}
\usepackage{lipsum}
\usepackage{enumitem}
\usepackage{graphicx}
\usepackage{tikz}
\usepackage{tikzducks}

\usepackage{xcolor}
\usepackage[table]{xcolor}

\usepackage{algorithm}
\usepackage{algorithmic}

\usepackage{amsmath}        % \text, \bigl/\bigr, align 등               
\usepackage{amssymb}  
\usepackage{mathtools}
\usepackage[labelfont=bf]{caption}
\usepackage{pifont}
\usepackage{tabularx}
\usepackage{hyperref}

\newcommand{\paragrapht}[1]{\noindent\textbf{#1}}

\definecolor{dark_blue}{rgb}{0, 0, 0.7}

\title{Controllable Dynamic 3D Shape Generation via \\3D Trajectories and Text}

\newcommand{\ours}{T2Mo}

\author{
  Jaeyeong Kim \quad 
  Ines Kim \quad 
  Jahyeok Koo \quad 
  Seungryong Kim\\[5pt]
  KAIST AI \\[5pt]
  \small \href{https://cvlab-kaist.github.io/T2Mo/}{\textbf{Project Page}: \texttt{https://cvlab-kaist.github.io/T2Mo}} \\
}
\begin{document}
\maketitle
\begin{figure}[h]
    \centering
    \includegraphics[width=\textwidth]{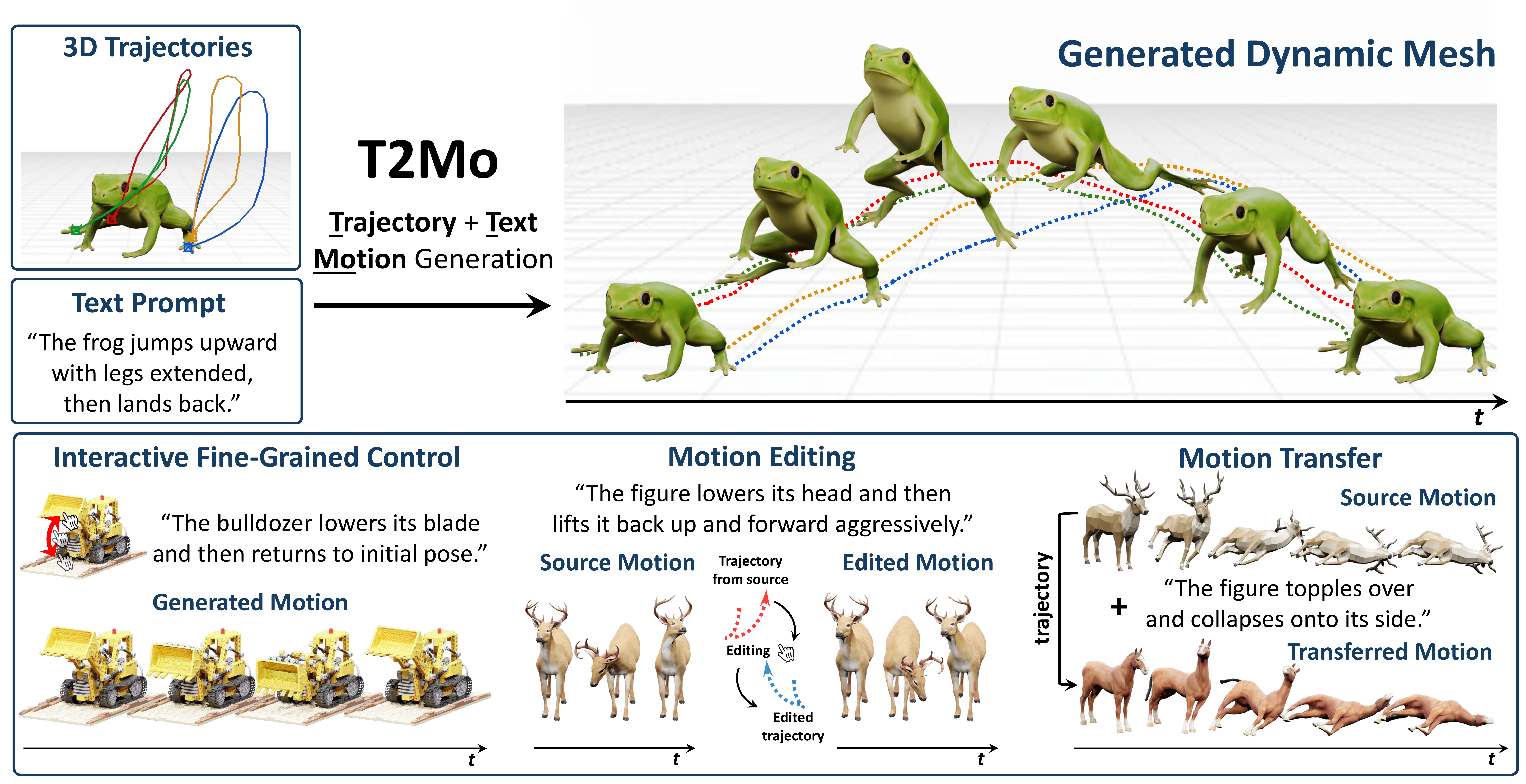} 
    \caption{
    We present \textbf{T2Mo}, a feed-forward framework for controllable dynamic 3D shape generation guided by 3D trajectories and text. By combining spatial trajectory conditioning with semantic text conditioning, T2Mo generates spatially controlled motions. The versatility of trajectories also enables diverse applications, including interactive fine-grained control, motion editing, and motion transfer.
    }
    \label{fig:teaser}
\end{figure}

\begin{abstract}
We introduce \textbf{T2Mo}, a feed-forward framework for controllable dynamic 3D shape generation conditioned on 3D trajectories and text. Due to the inherent ambiguity of language, generating precisely intended motions using text alone remains challenging. To address this, we adopt 3D trajectories as controllable spatial guidance, specifying the exact paths along which selected points should move. By combining both, T2Mo generates object motions that spatially adhere to the given trajectories while globally reflecting the text semantics. To robustly handle trajectory inputs with arbitrary configurations—ranging from dense to sparse and unevenly distributed—we further propose a shape-grounded trajectory embedding that maps an input trajectory set into a shape-aware token set covering the entire object. We conduct extensive comparisons against text-based baselines and cascaded video-based baselines that combine trajectory-guided video generation with video-to-dynamic mesh generation. Quantitative and qualitative evaluations, along with user studies, demonstrate that our approach produces motions that more faithfully follow the given prompts with higher expressiveness while preserving motion quality.
\end{abstract}

% % Previous version
% We present T2Mo, the first feed-forward framework for controllable dynamic 3D shape generation of general objects conditioned on 3D trajectories and text prompts. Existing text-driven methods struggle to specify precise motion due to the ambiguity of natural language. To address this, we introduce 3D trajectories as a direct spatial control signal and propose a shape-grounded trajectory embedding that maps arbitrary user-provided trajectories into geometry-aware condition tokens aligned with the input shape. Injecting these tokens into the generative backbone enables highly controllable motion generation while handling varying trajectory numbers and distributions. We conduct extensive comparisons against text-based baselines and trajectory-guided video generation workarounds. Quantitative and qualitative evaluations, along with user studies, show that our method produces motions that more faithfully follow the given prompts with higher expressiveness while preserving motion quality. %Our code and weights will be publicly released.

\section{Introduction}

Dynamic 3D shape generation has become an important research problem because of its wide‑ranging applications in AR/VR, gaming, filmmaking, and other domains that require temporally coherent 3D assets. Recent advances in feed‑forward 3D generative models~\cite{li2025triposg, li2025step1x, li2025craftsman3d, xiang2025structured, pmlr-v267-han25i, zhang20233dshape2vecset, lai2025hunyuan3d, zhang2024clay} have made it possible to synthesize dynamic 3D shapes more efficiently than earlier optimization‑based methods~\cite{bahmani20244d, li2025articulated, yu20244real, ling2024align} by extending 3D generative models in the temporal dimension~\cite{sabathier2026actionmesh, yenphraphai2025shapegen4d, jiang2026mesh4d} or generating motion of a given 3D object~\cite{Wu_2025_ICCV, wang2025bimotion, wu2026animateanymeshflexible4dfoundation, chen2026motion}. Despite these advances, controllability of dynamic shape generation remains underexplored, as current models lack explicit mechanisms to precisely specify the intended motion. Providing users with intuitive control over how objects move remains an open challenge, motivating the need for controllable dynamic 3D shape generation.

To achieve controllability in visual generation tasks, recent approaches employ text conditioning as high-level semantic guidance alongside additional control signals to mitigate the inherent ambiguity of language. For instance, in video synthesis and human motion generation, prior works introduce auxiliary controls such as bounding boxes~\cite{ma2024trailblazer, wang2024boximator}, motion vectors~\cite{yin2023dragnuwa, chu2025wan, Geng_2025_CVPR, Zhang_2025_CVPR, wang2025levitor}, segmentation masks~\cite{wu2024draganything}, or kinematic constraints~\cite{wan2024tlcontrol, kania2021trajevae, guo2025motionlab, rempe2026kimodo}. While these methods enable controllable generation by allowing users to specify desired motions in pixel space or via domain-specific modalities like skeletons, they are highly tailored to their respective domains and cannot be directly applied to dynamic 3D mesh generation. Consequently, providing intuitive motion control for dynamic 3D generation remains an open problem.

We address this gap by representing user control signals directly on the mesh through \emph{3D trajectories}. By encoding the path of a point on the target mesh in 3D space across frames, 3D trajectories provide an intuitive and interactive way to express motion control without requiring any additional transformations. Building on this representation, we introduce \textbf{\ours}, a feed-forward framework for controllable dynamic 3D generation of general objects. The framework conditions the generative model on a set of user-provided trajectories together with a text prompt that provides global motion semantics. This enables our framework to support controllable motion generation that adheres to the user-provided trajectories while remaining consistent with the global semantics of the text prompt.

However, user-provided trajectories may take diverse configurations in practice—from dense to sparse and unevenly distributed—depending on the target motion, the object geometry, and the application. To robustly generate plausible motions across such varying trajectory configurations, the model should process each trajectory adaptively with respect to the input shape, while remaining consistent regardless of how the input trajectories are configured. For example, a trajectory placed on the leg of a person with the prompt "walking" should affect only that leg part, whether it is the sole input or one among many. To this end, we design a \textit{shape-grounded trajectory embedding} that maps the input trajectory set into a predefined fixed-size, shape-aware token set covering the entire shape independently of the input trajectory configuration. Intuitively, this input trajectory-agnostic coverage of the entire shape gives consistent conditioning regardless of the input configuration, while attaching each trajectory to a shape-aware token localizes its effect in a geometry-aware manner.

% 여기에 어플
We evaluate T2Mo on a diverse set of objects and compare it against text-based baselines~\cite{Wu_2025_ICCV, wang2025bimotion} and cascaded baselines which combine trajectory-conditioned video generation and video-to-dynamic mesh generation~\cite{Zhang_2025_CVPR, sabathier2026actionmesh, chen2026motion} through quantitative metrics including VBench~\cite{huang2024vbench}, trajectory alignment, and motion magnitude~\cite{wu2026animateanymeshflexible4dfoundation}, alongside user study. The comparison indicates that introducing trajectory guidance improves both controllability and expressiveness by providing explicit motion cues, without compromising the overall quality of the generated motion. We further showcase qualitative results, demonstrating that our framework enables a diverse range of motion controls and applications.

In summary, our contributions are as follows:

\begin{itemize}
\item We present \textbf{\ours}, a feed-forward framework for controllable dynamic 3D shape generation of general objects, conditioned on \emph{3D trajectories} and a \emph{text} prompt.
\item We introduce a \textbf{shape-grounded trajectory embedding} that maps arbitrarily configured input trajectories into a predefined shape-aware token set covering the entire object, providing consistent conditioning regardless of configuration while preserving geometric locality.
\item Extensive experiments and user preference studies demonstrate that our framework enables various user-defined motion controls, improving both controllability and expressiveness over recent baselines without compromising the quality of the generated motion.
\end{itemize}

\section{Related work}
% \textcolor{red}{Adding more citations video generation, human motion synthesis - in progress}

% \textcolor{red}{Clarify existing methods rely on text and video-based model focus on reconstruction. (BiMotion)}

% \textcolor{red}{Clarify drag based editing methods(BiMotion)}

% \textcolor{red}{Pupetteer}
\paragrapht{Dynamic 3D generation.}
Similar to static 3D generation, early works on dynamic 3D generation predominantly adopt optimization-based formulations~\cite{bahmani20244d, li2025articulated, yu20244real, ling2024align, li2024dreammesh4d}. Such methods commonly distill a 2D generative prior through Score Distillation Sampling (SDS)~\cite{poole2022dreamfusion} and its variants~\cite{wang2023prolificdreamer}. 
However, due to their limitations such as view inconsistency and the Janus problem~\cite{hong2023debiasing}, subsequent works shifted toward extending 2D generative models to multi-view settings~\cite{jiang2024animate3d, qin2025distilling, xie2025videopanda, zhang20244diffusion} for dynamic 3D reconstruction~\cite{gao2025charactershot, huang2025mvtokenflow, huang2025animax, ren2024l4gm, sun2024eg4d, wu2025cat4d}. 
Recently, as feed-forward 3D shape generation models~\cite{li2025triposg, li2025step1x, li2025craftsman3d, xiang2025structured, pmlr-v267-han25i, zhang20233dshape2vecset, lai2025hunyuan3d, zhang2024clay} have shown remarkable progress, there have been attempts to leverage these models to extend static 3D generation into dynamic generation ~\cite{yenphraphai2025shapegen4d, sabathier2026actionmesh}.
Alternatively, an emerging line of work orthogonally decomposes this task, independently handling static 3D asset creation and per-vertex trajectory synthesis~\cite{Wu_2025_ICCV, chen2026motion, jiang2026mesh4d, wang2025bimotion}.
Building on this idea, recent text-guided dynamic 3D generation methods adopt the approach of animating predefined 3D meshes. 
Specifically, AnimateAnyMesh~\cite{Wu_2025_ICCV} and AnimateAnyMesh++~\cite{wu2026animateanymeshflexible4dfoundation} utilize a rectified flow model with trajectory VAE to generate motion conditioned on a text and an initial shape. 
BiMotion~\cite{wang2025bimotion} introduces a B-spline-based motion representation to handle motion of arbitrary frame lengths. 
Adopting this decoupling approach, we generate per-vertex displacements over time for a given input shape. This formulation is particularly suitable for controllable generation, as it allows users to directly provide guidance based on the underlying geometry.
Although these advancements successfully generate dynamic 3D shapes from given contexts in a feed-forward manner, they remain limited in providing the sufficient controllability required for practical applications. 
Motivated by this limitation, we propose a controllable generation approach for general objects through trajectory-based guidance.

% Video generation, one of the most representative task, where trajectories defined on the shared 2D pixel space are used to guide object or camera motion 

\paragrapht{Trajectory-guided motion control.}
Owing to their spatiotemporal expressiveness, point trajectories have been widely adopted as a versatile representation for controllable generation across a range of tasks. Video generation, one of the most representative tasks, commonly employs sparse or dense trajectories in 2D pixel space as a spatial guidance of object or camera motion. 
% Since the conditioning trajectories and the generation target reside in the same 2D pixel space, this inherent spatial correspondence can be exploited to inject the condition without additional explicit alignment.
Existing approaches integrate such trajectory guidance through feature-level modulation~\cite{chu2025wan}, ControlNet-style conditioning~\cite{Geng_2025_CVPR, wang2025levitor}, or specialized trajectory encoders~\cite{Zhang_2025_CVPR, yin2023dragnuwa, wu2024draganything, wang2024motionctrl}. 
More recent extensions move beyond 2D image-plane trajectories by introducing 3D pose sequence~\cite{fu20243dtrajmaster} or trajectory with depth~\cite{wang2025levitor}, mainly focusing on specifying the object-level global position trajectory.
Similar principles have been applied to human motion synthesis~\cite{kania2021trajevae, wan2024tlcontrol, guo2025motionlab, rempe2026kimodo, xie2024omnicontrol, karunratanakul2023guided}, utilizing trajectories as spatial constraints to guide global paths or specific actions, often conjugated with other condition modalities.
However, these methods are largely limited to articulated human body motion, where the control mechanism is tightly coupled to a predefined skeletal structure. 
In the context of 4D generation, although TC4D~\cite{10.1007/978-3-031-72952-2_4} proposes trajectory-conditioned generation, trajectories are used solely to define the scene-level per-object position paths, rather than to model local object motion itself, leaving the local, intra-object motion uncontrolled.
Unlike previous works, we propose a general-purpose framework that establishes 3D trajectories as a versatile interface for motion control.

\section{Method}
\subsection{Overview}
Our goal is to enable explicit control over dynamic 3D shape generation using two complementary signals: a \emph{set of 3D trajectories} that specify point-wise motion on the mesh and a \emph{text prompt} that conveys high-level motion semantics. However, in practice, user-provided trajectories are often sparse and unevenly distributed, and naively injecting them often fails to yield a well-localized conditioning signal on the shape. To handle such variability, we introduce \textbf{shape-grounded trajectory embedding}, which converts input trajectory set with arbitrary configuration into a predefined shape-aware token set covering the entire shape. These tokens, together with text embeddings, condition a diffusion-based generative model built on DiT~\cite{peebles2023scalable}, which synthesizes per-vertex displacements for a given static mesh. This section first formalizes the task (Sec.~\ref{method:Background}), then describes the shape-grounded trajectory embedding and generative architecture (Sec.~\ref{method:MainBody}), and outlines training and inference (Sec.~\ref{method:TrainInference}).

\begin{figure}[t!]
    \centering
    \includegraphics[width=1.0\textwidth]{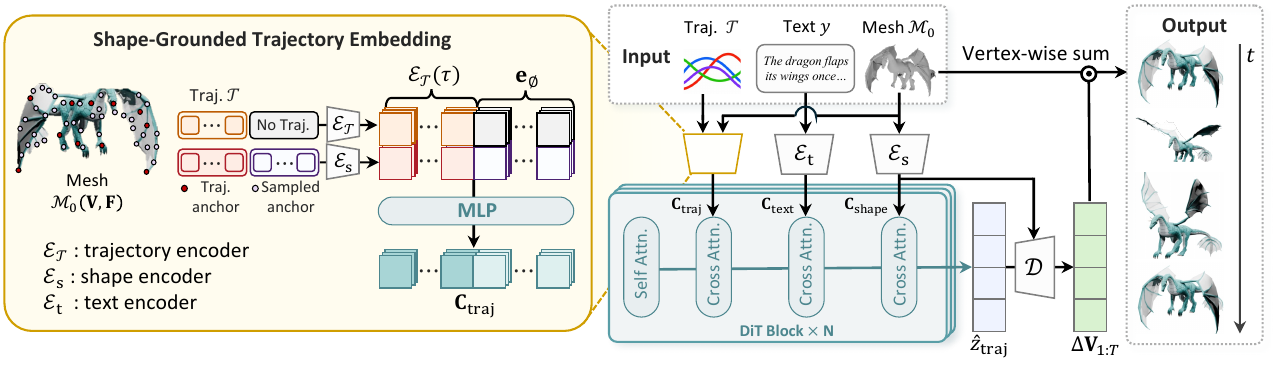} 
    \caption{
        \textbf{Overview of T2Mo.}
        Given a source mesh \(\mathcal{M}_{0}\), condition 3D trajectories \(\mathcal{T}\), and a text prompt \(y\), our model generates per-vertex displacements \(\Delta \mathbf{V}_{1:T}\) via a DiT~\cite{peebles2023scalable} backbone conditioned on shape \(\mathbf{C}_{\text{shape}}\), text \(\mathbf{C}_{\text{text}}\), and trajectory tokens \(\mathbf{C}_{\text{traj}}\) through cross-attention, yielding the output mesh sequence \(\{\mathcal{M}_{i}\}_{i=1}^{T}\) via the decoder \(\mathfrak{D}\). The trajectory condition is constructed by our \emph{shape-grounded trajectory embedding} (yellow box): each input trajectory is bound to its source point on the mesh (red dots), and the remaining anchors are sampled by farthest point sampling (white dots). For each anchor, the shape feature from \(\mathcal{E}_{\text{s}}\) is concatenated with either a trajectory feature from \(\mathcal{E}_{\mathcal{T}}\) or a learnable null embedding \(\mathbf{e}_{\varnothing}\), and projected through an MLP to form the trajectory conditions \(\mathbf{C}_{\text{traj}}\).
    }
    \vspace{-10pt}
    \label{fig:architecture}
\end{figure}

\subsection{Problem formulation} \label{method:Background}
We formalize controllable dynamic 3D shape generation as generating a sequence of deformed meshes given a static source mesh and user-provided controls. Let \(\mathcal{M}_0 = (\mathbf{V}_0, \mathbf{F}_0)\) be a source mesh with \(N\) vertices \(\mathbf{V}_0 \in \mathbb{R}^{N\times 3}\) and \(M\) faces \(\mathbf{F}_0 \in \mathbb{Z}^{M\times 3}\). The task is to generate per‑vertex displacements \(\Delta \mathbf{V}_{1:T} = \{\Delta \mathbf{V}_t\}_{t=1}^T\), where each \(\Delta \mathbf{V}_t \in \mathbb{R}^{N\times 3}\), over \(T\) frames. Each output frame is reconstructed as \(\mathcal{M}_t = (\mathbf{V}_0 + \Delta \mathbf{V}_t, \mathbf{F}_0)\). The user provides a text prompt \(y\) describing global motion semantics and a set of 3D trajectories \(\mathcal{T} = \{\tau_k\}_{k=1}^K\) of arbitrary cardinality \(K\). Each trajectory \(\tau_k = (\mathbf{p}_{k,0}, \mathbf{p}_{k,1}, \dots)\) is a sequence of 3D waypoints, where \(\mathbf{p}_{k,i} \in \mathbb{R}^3\) denotes the \(i\)-th waypoint.

Following prior work~\cite{wu2026animateanymeshflexible4dfoundation, Wu_2025_ICCV, wang2025bimotion}, we compress the static mesh and its motion into latent codes using a variational autoencoder (VAE). 
The VAE encoders map the source mesh and displacement sequence into a shape \(\mathbf{z}_{\text{shape}}\) and a trajectory latent \(\mathbf{z}_{\text{traj}}\) with two separate streams in the VAE encoder:
\begin{equation}
  \mathbf{z}_{\text{shape}} = \mathcal{E}_{\text{shape}}(\mathcal{M}_{0}),\qquad \mathbf{z}_{\text{traj}} = \mathcal{E}_{\text{traj}}(\Delta \mathbf{V}_{1:T})
\end{equation}
where \(\mathcal{E}_{\text{shape}}\) and \(\mathcal{E}_{\text{traj}}\) denote the shape and trajectory streams, respectively.
At inference, a generative model \(\mathcal{G}_\theta\) produces a trajectory latent \(\hat{\mathbf{z}}_{\text{traj}}\) from Gaussian noise \(\boldsymbol{\epsilon} \sim \mathcal{N}(\mathbf{0}, \mathbf{I})\), conditioned on \(\mathbf{z}_{\text{shape}}\) and the text and trajectory condition tokens \(\mathbf{C}_{\text{text}}\) and \(\mathbf{C}_{\text{traj}}\):
\begin{equation}
  \hat{\mathbf{z}}_{\text{traj}} = \mathcal{G}_{\theta}(\boldsymbol{\epsilon};\, \mathbf{z}_{\text{shape}},\, \mathbf{C}_{\text{text}}, \mathbf{C}_{\text{traj}}),\quad \boldsymbol{\epsilon} \sim \mathcal{N}(\mathbf{0}, \mathbf{I}),\qquad
  \Delta \mathbf{V}_{1:T} = \mathcal{D}(\mathbf{z}_{\text{shape}},\, \hat{\mathbf{z}}_{\text{traj}}),
\end{equation}
where \(\mathcal{D}\) is the VAE decoder. For notational consistency, we hereafter define
\(\mathbf{C}_{\text{shape}} \coloneqq \mathbf{z}_{\text{shape}}\).

\subsection{\ours: Trajectory and text-guided dynamic 3D generation} \label{method:MainBody}
Fig.~\ref{fig:architecture} illustrates the overall architecture of \textbf{\ours}, including the generative model and the shape‑grounded trajectory embedding.

\paragrapht{Shape‑grounded trajectory embedding.} 
Simply treating each trajectory as an independent token neglects the geometry of the shape and leads to poorly localized conditioning (Fig. ~\ref{fig:traj-enc-abl-qual}). 
Instead, we sample a fixed set of \(N_{\text{anchor}}\) anchor points \(\{(\mathbf{s}_i, \mathbf{n}_i)\}_{i=1}^{N_{\text{anchor}}}\) on the surface, where each anchor consists of a position \(\mathbf{s}_i \in \mathbb{R}^{3}\) and a surface normal \(\mathbf{n}_i \in \mathbb{R}^{3}\). Specifically, we first place the source points \(\{\mathbf{p}_{k, 0}\}_{k=1}^{|\mathcal{T}|}\) of the given trajectory set \(\mathcal{T}\) as \emph{trajectory anchors}. We then fill the remaining anchor points with \emph{sampled anchors} obtained by farthest point sampling seeded with the trajectory anchors. By construction, anchor \(i \leq |\mathcal{T}|\) corresponds to trajectory \(\tau_i\). For each anchor \(i\), we obtain a per-anchor embedding \(\mathbf{e}_i\) from a trajectory embedding \(\mathcal{E}_{\text{traj}}(\tau_i)\) when \(i\) is a trajectory anchor (\(i \leq |\mathcal{T}|\)), and from a learnable null embedding \(\mathbf{e}_\varnothing\) otherwise. The conditioning token \(\mathbf{c}_i\) for each anchor is then formed by concatenating the shape encoder's output \(\mathcal{E}_{\text{shape}}(\mathbf{s}_i, \mathbf{n}_i)\) with \(\mathbf{e}_i\), followed by an MLP:
\begin{equation}
  \mathbf{c}_{i} = \mathrm{MLP}\!\left(\,\mathrm{concat}\!\left(\mathcal{E}_{\text{shape}}(\mathbf{s}_{i}, \mathbf{n}_{i}),\; \mathbf{e}_{i}\right)\right),
  \quad \text{where} \quad \mathbf{e}_{i} =
  \begin{cases}
    \mathcal{E}_{\text{traj}}(\tau_{i}) & \text{if } i \leq |\mathcal{T}|, \\
    \mathbf{e}_{\varnothing} & \text{otherwise}.
  \end{cases}
  \label{equ:shape-grounded_trajectory_embedding}
\end{equation}
% The resulting matrix \(\mathbf{C}_{\text{traj}}\in \mathbb{R}^{N_{\text{anchor}}\times D}\) provides dense, geometry‑aware conditioning tokens that cover the entire shape and indicate which regions are guided by trajectories.
The resulting matrix \(\mathbf{C}_{\text{traj}} \in \mathbb{R}^{N_{\text{anchor}}\times D}\) with feature dimension \(D\) provides consistent geometry-aware conditioning tokens that cover the entire shape and indicate which regions are guided by trajectories.

\paragrapht{Generative architecture.} Our generative model builds on DiT~\cite{peebles2023scalable} and employs cross‑attention to integrate the shape, trajectory and text conditions. For each DiT block, we compute keys and values from the shape tokens \(\mathbf{C}_{\text{shape}}\) (via \(\mathcal{E}_{\text{shape}}\)), the trajectory tokens \(\mathbf{C}_{\text{traj}}\) (via shape-grounded trajectory embedding), and text tokens \(\mathbf{C}_{\text{text}}\) (via a CLIP text encoder~\cite{pmlr-v139-radford21a}). 
We initialize the text and shape branches with weights from~\cite{wang2025bimotion} and add the trajectory branch through a gated cross‑attention~\cite{alayrac2022flamingo} to smoothly incorporate trajectory guidance:
\begin{equation}
\begin{gathered}
    \mathbf{h} \leftarrow \mathbf{h} + \tanh(\alpha)\cdot\mathrm{Attn}(
    \mathbf{Q}_{\mathbf{h}},\, \mathbf{K}_{\text{traj}},\,
    \mathbf{V}_{\text{traj}}), \\
    \mathbf{Q}_{\mathbf{h}} = \mathbf{h}\mathbf{W}_{q},\quad
    \mathbf{K}_{\text{traj}} = \mathbf{C}_{\text{traj}}\mathbf{W}_{k},\quad
    \mathbf{V}_{\text{traj}} = \mathbf{C}_{\text{traj}}\mathbf{W}_{v},
\end{gathered}
\end{equation}
where \(\mathbf{h}\) denotes the hidden state and \(\alpha \in \mathbb{R}\) is a learnable scalar gate initialized to zero, so the backbone initially ignores trajectory input.

\subsection{Training and inference pipeline} \label{method:TrainInference}

\paragraph{Training.} 
We train the model in two stages, inspired by the conditional diffusion training schemes in prior work~\cite{Zhang_2025_CVPR, yin2023dragnuwa, wang2024motionctrl}. Stage one conditions on trajectories assigned to all \(N_{\text{anchor}}\) anchors, teaching the model to use dense guidance. Stage two gradually sparsifies the guidance by randomly sampling \(n\) trajectories per sample, with the range of \(n\) annealed from \(N_{\text{anchor}}\) down to a minimum value. We train the generative model using a rectified flow objective~\cite{esser2024scaling}. Given a noisy latent \(\hat{\mathbf{z}}_s = (1-s)\,\mathbf{z}_0 + s\,\mathbf{z}_1\) interpolated from random noise \(\mathbf{z}_0\sim \mathcal{N}(\mathbf{0},\mathbf{I})\) and a clean latent \(\mathbf{z}_1\) with a scalar \(s\sim \mathcal{U}(0,1)\), the model predicts the velocity \(\mathbf{v}_s = \mathbf{z}_1 - \mathbf{z}_0\) and minimizes:
\begin{equation}
  \mathcal{L} = \mathbb{E}_{\mathbf{z}_{0}, \mathbf{z}_{1}, s, \mathbf{C}}\!\left[\,\big\| \mathbf{v}_{\theta}(\hat{\mathbf{z}}_{s},\, s,\, \mathbf{C}) - (\mathbf{z}_{1} - \mathbf{z}_{0}) \big\|_{2}^{2}\,\right],
\end{equation}
where \(\mathbf{C} = \{\mathbf{C}_{\text{traj}}, \mathbf{C}_{\text{text}}, \mathbf{C}_{\text{shape}}\}\) denotes the conditioning tokens defined in Sec.~\ref{method:MainBody}.

\paragraph{Inference.} 
At inference time, we solve the deterministic ODE associated with rectified flow. We apply classifier-free guidance~\cite{ho2022classifier} only to the text branch. Given the velocity fields predicted with only the trajectory tokens (\(\mathbf{v}_\theta^{\text{traj}}\)) and with both the text and trajectory tokens (\(\mathbf{v}_\theta^{\text{both}}\)), we compute the guided velocity as:
\begin{equation}
    \mathbf{v}_{\theta}^{\text{cfg}} = \mathbf{v}_{\theta}^{\text{traj}} + \gamma \cdot (\mathbf{v}_{\theta}^{\text{both}} - \mathbf{v}_{\theta}^{\text{traj}}),
\end{equation}
with a guidance scale \(\gamma\). 
We do not guide the trajectory branch because the trajectory already explicitly provides its magnitude, and amplifying it tends to cause exaggerated motions. Users can specify arbitrary trajectories, which we discretize into waypoint sequences before passing them to the model.

\begin{table}[t!]
\centering
\small
\setlength{\tabcolsep}{4pt}
\caption{
\textbf{Quantitative comparison with text-conditioned baselines.} The \emph{Cond.} column indicates the type of input conditions used by each method (text or trajectory). Our method outperforms the baselines~\cite{Wu_2025_ICCV, wang2025bimotion} across the user study, most VBench~\cite{huang2024vbench} metrics, and AMD~\cite{wu2026animateanymeshflexible4dfoundation}, while remaining competitive on motion smoothness (MS).
% By leveraging trajectory guidance, our method achieves better expressiveness and prompt alignment, while preserving motion quality.
}
\label{tab:quan_w_text}

% 우측 잘림 방지를 위해 \columnwidth에 맞게 리사이즈합니다.
\resizebox{0.9\columnwidth}{!}{%
\begin{tabular}{l cc ccc cccc c}
\toprule
\multirow{2}{*}{\textbf{Method}} 
& \multicolumn{2}{c}{\textbf{Cond.}} 
& \multicolumn{3}{c}{\textbf{User Study} $\uparrow$} 
& \multicolumn{4}{c}{\textbf{VBench} $\uparrow$}
& \multicolumn{1}{c}{\textbf{Motion Mag.} $\uparrow$} \\
\cmidrule(r){2-3} \cmidrule(lr){4-6} \cmidrule(lr){7-10} \cmidrule(l){11-11}
& Text & Traj & PA & MP & ME & OC & AQ & MS & DD & AMD \\
\midrule
\midrule
AnimateAnyMesh~\cite{Wu_2025_ICCV} & \ding{51} & \ding{55} & 2.170 & 2.667 & 2.560 & 0.168 & 0.539 & \textbf{0.992} & 0.585 & 0.043 \\
BiMotion~\cite{wang2025bimotion}   & \ding{51} & \ding{55} & 2.657 & 3.040 & 2.813 & 0.170 & 0.536 & 0.991 & 0.329 & 0.046 \\
\midrule
% \textbf{Ours} w/o trajectory & \ding{51} & \ding{55} & \textbf{-} & \textbf{-} & \textbf{-} & \textbf{-} & \textbf{-} & \textbf{-} & \textbf{-} & \textbf{-} \\
% \textbf{Ours} w/ 1 trajectory  & \ding{51} & \ding{51} & \textbf{-} & \textbf{-} & \textbf{-} & \textbf{0.18} & \textbf{0.551} & \textbf{0.98} & \textbf{0.51} & \textbf{-} \\
\textbf{Ours}  & \ding{51} & \ding{51} & \textbf{4.507} & \textbf{4.067} & \textbf{4.340} & \textbf{0.175} & \textbf{0.544} & 0.984 & \textbf{0.743} & \textbf{0.147} \\
\bottomrule
\end{tabular}%
}
\end{table}

\section{Experiments}

\subsection{Experimental setup}
\paragraph{Implementation details.}
We train a rectified flow-based generative model on the BiMO dataset, building on the pretrained VAE of~\cite{wang2025bimotion}. Our model consists of 12 DiT blocks and is trained for 100K iterations at a learning rate of \(5\times 10^{-5}\), using the Adam optimizer~\cite{kingma2014adam} with weight decay \(1\times 10^{-2}\). We linearly warm up the learning rate over the first 1\(K\) iterations and then decay it to \(1\times 10^{-5}\) via scheduled annealing. Training is performed on four NVIDIA RTX 3090 GPUs in 3 days. For each object, we randomly sample 4096 points from the source mesh and further select 512 of them via farthest point sampling to be encoded into latent tokens by the VAE encoder. For the shape-grounded trajectory embedding, we use \(N_{\text{anchor}}\!= 64\) anchor points. The text condition is encoded with CLIP ViT-L/14~\cite{pmlr-v139-radford21a}, using a maximum token length of 77. For sampling, we use \(\gamma=3.0\) for classifier-free guidance~\cite{ho2022classifier}. To improve robustness against noisy and coarse input trajectories, we perturb each conditioning trajectory with random Gaussian noise, and apply two smoothing techniques: 1D Gaussian filtering and trajectory simplification. Further augmentation details are provided in Appendix~\ref{appendix:data-aug}.

\paragraph{Baselines.} 
We compare our model against text-conditioned baselines, AnimateAnyMesh~\cite{Wu_2025_ICCV} and BiMotion~\cite{wang2025bimotion}. 
To further assess controllability, we additionally consider video-based baselines that incorporate trajectory conditioning indirectly by combining trajectory-guided video generation with video-based dynamic 3D generation. 
Specifically, we adopt Tora~\cite{Zhang_2025_CVPR}, a trajectory-conditioned video generator based on CogVideoX 5B~\cite{yang2024cogvideox} that accepts trajectories at arbitrary granularity, and pair it with two video-to-dynamic mesh generation models, ActionMesh~\cite{sabathier2026actionmesh} and Motion 3-to-4~\cite{chen2026motion}.

\paragraph{Evaluation protocol.}
Since BiMO~\cite{wang2025bimotion} provides no evaluation set for dynamic meshes, we conduct quantitative evaluation on 30 animated meshes sourced from Objaverse~\cite{deitke2023objaverse} and Sketchfab, a scale consistent with prior works~\cite{Wu_2025_ICCV, wang2025bimotion}, which use 10 and 20 meshes, respectively. We assess the generated dynamic meshes along three perspectives: (i) \emph{motion quality}, the visual fidelity of the generation; (ii) \emph{motion expressiveness}, how dynamic the generated motion is; and (iii) \emph{prompt alignment}, how faithfully the model follows the given text or trajectory condition. Following~\cite{jiang2024animate3d, Wu_2025_ICCV, wang2025bimotion, wu2026animateanymeshflexible4dfoundation}, we evaluate these perspectives with VBench~\cite{huang2024vbench}, reporting four sub-metrics: \emph{aesthetic quality} (AQ) and \emph{motion smoothness} (MS) for quality, \emph{dynamic degree} (DD) for expressiveness, and \emph{overall consistency} (OC) for prompt alignment. We additionally report \emph{Average Motion Distance} (AMD)~\cite{wu2026animateanymeshflexible4dfoundation}, which captures motion magnitude as the average inter-frame vertex displacement. We conduct a user study with 30 participants who rate the animations along three axes—\emph{prompt alignment} (PA), \emph{motion plausibility} (MP), and \emph{motion expressiveness} (ME)—on a 5-point Likert scale ranging from 1 (\emph{very poor}) to 5 (\emph{excellent}) on 10 randomly chosen samples per baseline type, text-conditioned and video-generation-based baselines. For trajectory-conditioned baselines, we additionally report \emph{Average Displacement Error} (ADE) and \emph{Final Displacement Error} (FDE), which are the average \(\ell_{2}\) distance between each input trajectory and its corresponding generated vertices, taken over all frames and at the final frame, respectively. We measure these metrics in the 2D pixel space and 3D space for video-based baselines. All evaluations are run on 16-frame videos rendered from fixed viewpoints; for VBench, we average the metrics over videos from two side viewpoints to mitigate viewpoint-dependent variance such as occlusion. Importantly, to reflect realistic settings where dense trajectories are unlikely to be provided, all quantitative evaluations use sparse trajectory sets of at most five trajectories per object; to rule out cherry-picking, these are sampled by a predefined rule-based procedure without per-instance selection or human intervention. For qualitative examples, only the trajectories shown in the figures are used. Additional details are provided in Appendix~\ref{appendix:eval-detail}.

\begin{figure}[t]
    \centering
    \includegraphics[width=\textwidth]{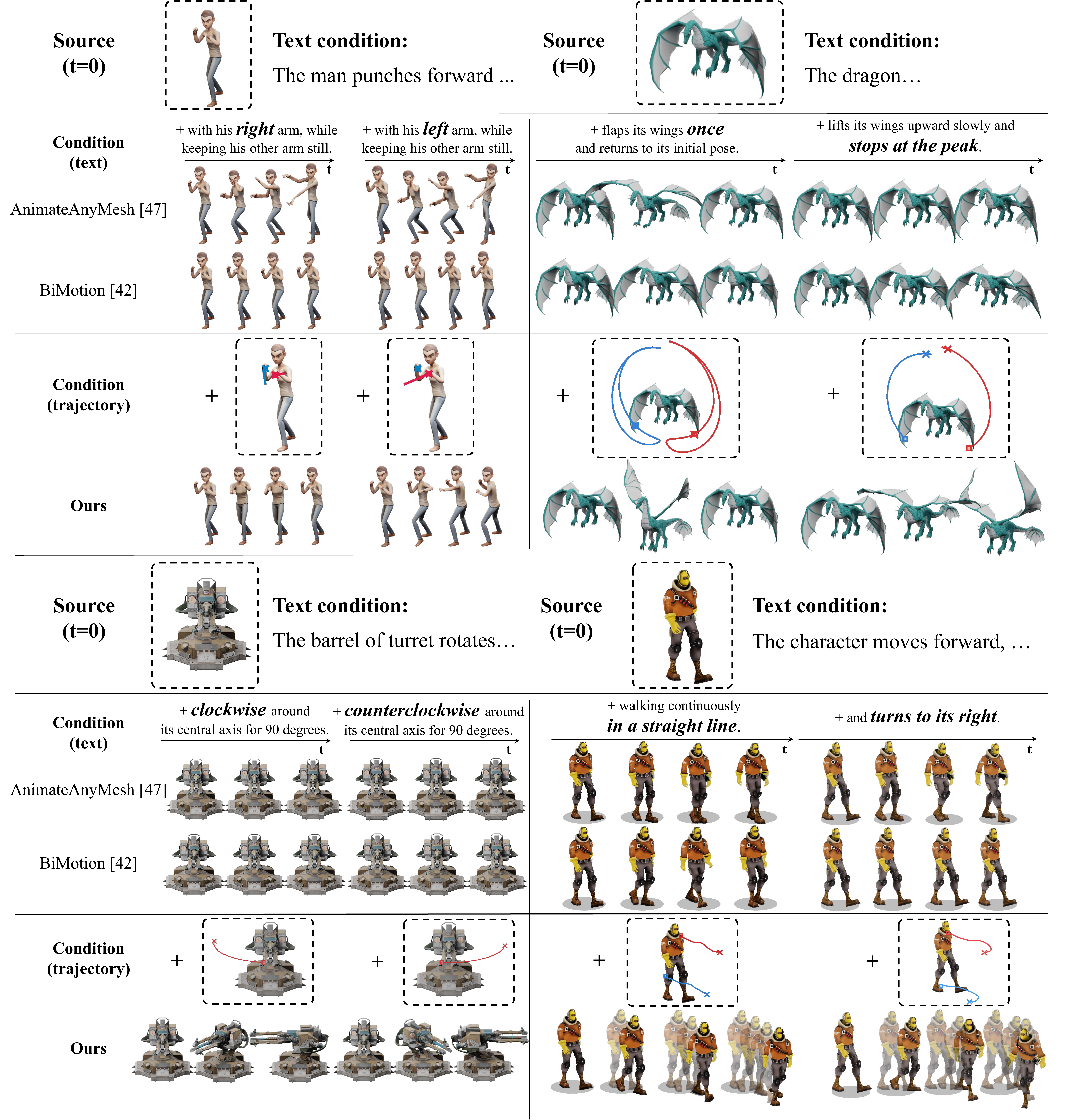}

    % \begin{tikzpicture}
        % \duck[water=cyan!50, magical=purple!50, laughing]
    % \end{tikzpicture}
    \caption{
    \textbf{Qualitative comparison with text-based baselines.}
    The baselines~\cite{Wu_2025_ICCV, wang2025bimotion} are conditioned on a pair of text prompts describing different specific motion details, while our model is guided by sparse trajectories representing those details. Our model uses only the visualized trajectories as inputs—at most two per example. By providing direct spatial cues, our method faithfully generates the intended motion, whereas the text-conditioned baselines fail to produce expressive motions aligned with the text and show little meaningful difference across the varying prompts. We provide video results in our supplementary materials.
    }
    \label{fig:qual-comparison-w-text}\vspace{-10pt}
\end{figure}

\begin{table*}[t!]
\centering
\scriptsize
\caption{
\textbf{Quantitative comparison with cascaded video-based baselines.} Our method achieves substantially lower trajectory error in both 2D pixel and 3D spaces, and outperforms the cascaded baselines across the user study, most VBench~\cite{huang2024vbench} metrics, and AMD~\cite{wu2026animateanymeshflexible4dfoundation}.
}
\label{tab:quan_w_video}

% tabularx 대신 tabular를 쓰고 resizebox로 가로폭(\textwidth)에 딱 맞춥니다.
\resizebox{\textwidth}{!}{%
\begin{tabular}{
  l
  cccc % Traj. Error (3D ADE, 3D FDE, 2D ADE, 2D FDE)
  ccc % User Study (PA, MP, ME)
  cccc % VBench (OC, AQ, MS, DD)
  c % Motion Magnitude (AMD)
}
\toprule
\multirow{2}{*}{\textbf{Method}}
& \multicolumn{4}{c}{\textbf{Traj. Error} $\downarrow$}
& \multicolumn{3}{c}{\textbf{User Study} $\uparrow$}
& \multicolumn{4}{c}{\textbf{VBench} $\uparrow$}
& \multicolumn{1}{c}{\textbf{Motion Mag.} $\uparrow$} \\
\cmidrule(lr){2-5}
\cmidrule(lr){6-8}
\cmidrule(lr){9-12}
\cmidrule(lr){13-13}
& ADE$_{\text{3D}}$ & FDE$_{\text{3D}}$ & ADE$_{\text{2D}}$ & FDE$_{\text{2D}}$
& PA & MP & ME
& OC & AQ & MS & DD
& AMD \\
% \midrule

% GPT5~\cite{singh2025openai} + AnimateAnyMesh~\cite{Wu_2025_ICCV}
% & - & - & - & - & - & - & - & - & - & - & - & - \\

% GPT5~\cite{singh2025openai} + BiMotion~\cite{wang2025bimotion}
% & - & - & - & - & - & - & - & - & - & - & - & - \\

\midrule

Tora~\cite{Zhang_2025_CVPR} + ActionMesh~\cite{sabathier2026actionmesh}
& 0.791 & 0.734 & 92.55 & 83.52
& 1.780 & 1.687 & 2.040 
& 0.171 & 0.506 & 0.988 & 0.486 
& 0.052 \\

Tora~\cite{Zhang_2025_CVPR} + Motion 3-to-4~\cite{chen2026motion}
& 0.605 & 0.675 & 69.19 & 77.64 
& 1.940 & 1.800 & 1.903 
& 0.166 & 0.522 & \textbf{0.991} & 0.486 
& 0.042 \\

\midrule

\textbf{Ours}
& \textbf{0.089} & \textbf{0.071} & \textbf{12.52} & \textbf{8.52}
& \textbf{4.577} & \textbf{4.177} & \textbf{4.437} 
& \textbf{0.175} & \textbf{0.544} & 0.984 & \textbf{0.743} 
& \textbf{0.147} \\

\bottomrule
\end{tabular}%
} % resizebox 끝
\end{table*}
\subsection{Comparisons}
\paragraph{Quantitative comparison with text-based baselines.} Tab.~\ref{tab:quan_w_text} summarizes the quantitative comparison between our model and the text-based generation baselines~\cite{wang2025bimotion, Wu_2025_ICCV}. 
Our method outperforms these baselines across the user study, most VBench~\cite{huang2024vbench} metrics, and AMD~\cite{wu2026animateanymeshflexible4dfoundation}. 
These results indicate that, by providing direct cues for the intended motion through the input trajectories, motion generated by our method follows the text prompt (PA, OC) more faithfully and produces higher quality motions (MP, AQ) and expressive motion (ME, DD, AMD) than the baselines. 
Although the baselines achieve slightly higher motion smoothness (MS) scores, MS rewards small frame-to-frame changes; together with the low ME in user study, AMD and dynamic degree (DD) scores of the baselines, this suggests that their MS scores stem from near-static or insufficiently dynamic motions.

\paragraph{Quantitative comparison with video-based baselines.} Tab.~\ref{tab:quan_w_video} reports the comparison with baselines that we construct by combining controllable video generation~\cite{Zhang_2025_CVPR} with video-based dynamic 3D generation~\cite{sabathier2026actionmesh, chen2026motion}. Across the VBench~\cite{huang2024vbench} and AMD~\cite{wu2026animateanymeshflexible4dfoundation}, these baselines fail to generate motion of sufficient quality and expressiveness aligned with the trajectory and text prompt, similar to the trend observed in the text-based baselines. 
The user study results show an even larger gap in motion quality and expressiveness, indicating that the video-based workaround often fails to produce high-quality text-conditioned dynamic 3D shape generation. 
The trajectory error metrics (ADE and FDE), in both the 2D pixel and 3D spaces, are roughly 5-10\(\times\) higher than those of our model, indicating that the control signal is not faithfully reflected.
Compared to both types of baselines, our method produces more expressive motion while preserving quality and prompt alignment.

\begin{figure}[t!]
    \centering
    \includegraphics[width=\textwidth]{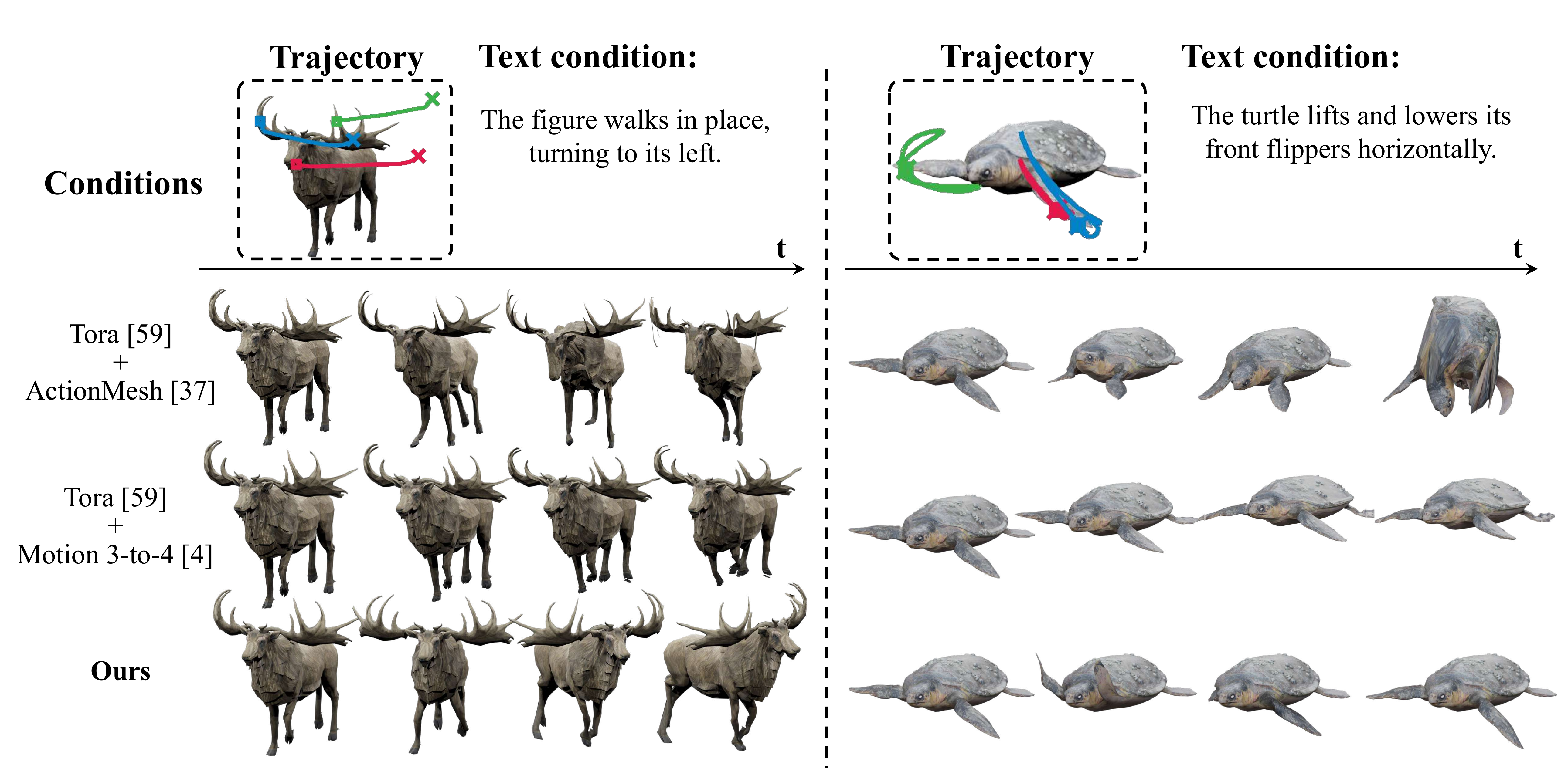}
    \caption{
    % \textcolor{red}{Video based control. Trajectory and detail will be changed.}
    % \textbf{Qualitative comparison with cascaded video-based baselines.} We compare our method against cascaded baselines~\cite{sabathier2026actionmesh, chen2026motion, Zhang_2025_CVPR} that rely on trajectory-conditioned video generation as a proxy. These baselines take 2D trajectories corresponding to the given camera view as input. The cascaded pipelines fail to synthesize dynamic 3D shapes that faithfully follow the trajectory guidance, and also struggle to preserve the visual structure of the input shape during motion.
    \textbf{Qualitative comparison with video-based baselines.} We compare our method against baselines that we construct by combining controllable video generation~\cite{Zhang_2025_CVPR} with video-based dynamic 3D generation~\cite{sabathier2026actionmesh, chen2026motion}. These baselines take 2D trajectories projected onto the given camera view as input. The constructed pipelines fail to synthesize dynamic 3D shapes that faithfully follow the trajectory guidance and struggle to preserve the visual structure of the input shape during motion.
    }
    \label{fig:qual-comparison-w-video}\vspace{-10pt}
\end{figure}
\begin{figure}[t!]
    \centering
    \includegraphics[width=\textwidth]{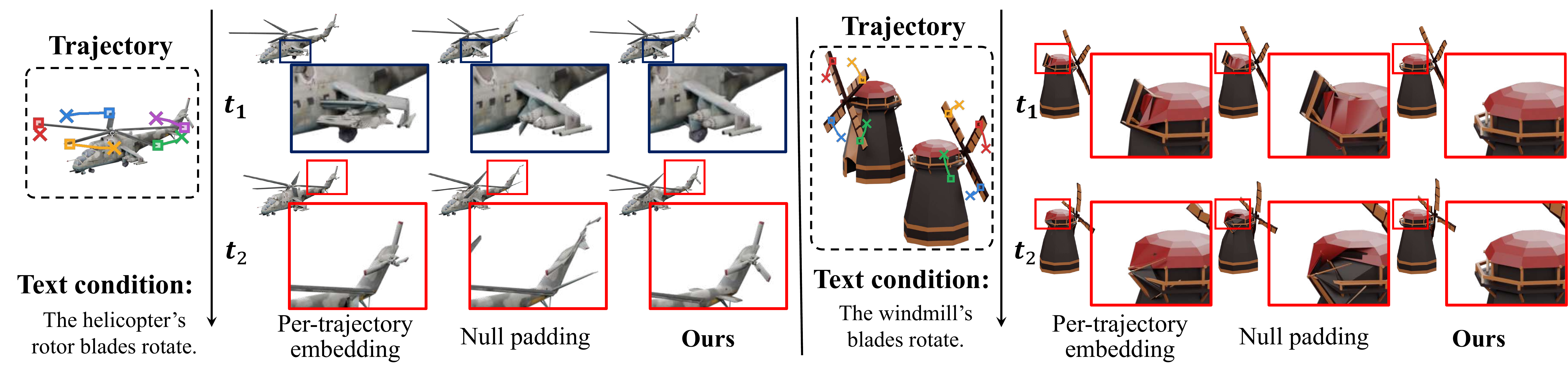}
    \caption{
    % \textcolor{red}{Trajectory encoder ablation. Trajectory and detail will be changed.} If trajectory is occluded at first frame, excluded that trajectory.
    \textbf{Qualitative comparison of trajectory embedding methods.} We compare our shape-grounded trajectory embedding against two naive variants: per-trajectory embedding and null-padded per-trajectory embedding. With the naive variants, the trajectory conditions leak into unintended regions and produce noticeable artifacts, losing geometric locality. For visualization clarity, only the first 10\% part of each trajectory is displayed.
    }
    \label{fig:traj-enc-abl-qual}\vspace{-10pt}
\end{figure}

% \paragrapht{Qualitative results.}
\paragraph{Qualitative comparison with text-based baselines.} 
We present qualitative results in Fig.~\ref{fig:qual-comparison-w-text} to compare the controllability, motion quality, and expressiveness of our generated motions against existing methods. To demonstrate the superior controllability of our trajectory-based approach over text-conditioned baselines~\cite{Wu_2025_ICCV, wang2025bimotion}, we task the models with generating specific motions: the text-based baselines are provided with detailed text descriptions, while our model is guided by sparse trajectories. As shown in the Fig.~\ref{fig:qual-comparison-w-text}, while the text-conditioned baselines struggle to realize finely specified motions from text prompts alone, our method successfully produces the intended motion by leveraging direct spatial guidance. This includes detail specification (top left), temporal control (top right), directional guidance (bottom left), and scene-level trajectory guidance (bottom right). 
% This result demonstrates the effectiveness of our trajectory-based spatial-guidance approach for precisely describing the intended motion.

% We present qualitative results in Fig.~\ref{fig:qual-comparison-w-text} to compare the controllability along with the quality and expressiveness of generated motion.
% To demonstrate the controllability of our trajectory-based approach compared to text-baselines~\cite{Wu_2025_ICCV, wang2025bimotion}, we generate specific motion by providing detailed text description pair for text-based baselines and sparse trajectory for our model.
% [trainsition?] While the text-conditioned baselines fail to realize finely specified motion by additional text prompt, our method successfully produces the intended motion by leveraging direct spatial guidance from the input trajectories—including detail specification (top left), temporal control (top right), directional control (bottom left), and global trajectory control (bottom right). 
% This result demonstrates the effectiveness of our trajectory-based spatial-guidance approach for precisely describing the intended motion.

\paragraph{Qualitative comparison with video-based baselines.} We further compare with the video-based baselines~\cite{Zhang_2025_CVPR, sabathier2026actionmesh, chen2026motion} in Fig.~\ref{fig:qual-comparison-w-video}. 
Our method generates motion that reflects both the trajectory and text conditions, whereas the combined baselines neither preserve the visual quality of the input shape nor faithfully follow the given trajectory prompt.

\begin{table*}[ht] % 전체 너비를 사용하는 table* 환경
    \centering
    
    % 왼쪽 칼럼용 minipage (Traj. Error 2개 + VBench 4개)
    \begin{minipage}{0.47\textwidth}
        \centering
        \caption{
        \textbf{Ablation study on the trajectory embedding.} \textbf{Top}: comparison of trajectory embedding variants. \textbf{Bottom}: effect of the number of anchor points \(N_{\text{anchor}}\).
        }
        \label{tab:ablation_embedding}
        \resizebox{\linewidth}{!}{%
        \begin{tabular}{l cc cccc}
            \toprule
            \multirow{2}{*}{\textbf{Method}} 
            & \multicolumn{2}{c}{\textbf{Traj. Error} $\downarrow$}
            & \multicolumn{4}{c}{\textbf{VBench} $\uparrow$} \\
            \cmidrule(lr){2-3} \cmidrule(lr){4-7}
            & ADE\(_{3\text{D}}\) & FDE\(_{3\text{D}}\) & OC & AQ & MS & DD \\
            
            \midrule
            \multicolumn{7}{l}{\textbf{\textit{Trajectory embedding strategy}}} \\
            \midrule
            Per-traj. emb. & 0.159 & 0.124 & 0.173 & 0.541 & \textbf{0.985} & \textbf{0.771} \\
            Null padding   & 0.101 & 0.080 & 0.174 & 0.540 & 0.984 & 0.743 \\
            Ours ($N_{\text{anchor}}=64$) & \textbf{0.089} & \textbf{0.071} & \textbf{0.175} & \textbf{0.544} & 0.984 & 0.743 \\
            
            \midrule
            \multicolumn{7}{l}{\textbf{\textit{Number of anchor points for training}}} \\
            \midrule
            Ours ($N_{\text{anchor}}=32$)  & 0.091 & \textbf{0.070} & \textbf{0.176} & \textbf{0.548} & \textbf{0.984} & 0.714 \\
            Ours ($N_{\text{anchor}}=64$)  & \textbf{0.089} & 0.071 & 0.175 & 0.544 & \textbf{0.984} & \textbf{0.743} \\
            Ours ($N_{\text{anchor}}=128$) & 0.092 & 0.074 & 0.175 & 0.544 & \textbf{0.984} & 0.714 \\
            Ours ($N_{\text{anchor}}=256$) & 0.090 & 0.071 & \textbf{0.176} & 0.543 & \textbf{0.984} & \textbf{0.743} \\
            
            % \midrule
            % \multicolumn{7}{l}{\textbf{\textit{Number of condition trajectories for inference}}} \\
            % \midrule
            % Ours ($N_{\text{cond}}=32$) & - & - & - & - & - & - \\
            % Ours ($N_{\text{cond}}=16$) &  &  & - & - & - & - \\
            % Ours ($N_{\text{cond}}<5$)  & - & - & - & - & - & - \\
            \bottomrule
        \end{tabular}
        }
    \end{minipage}
    \hfill % 왼쪽과 오른쪽 사이를 띄움
    % 오른쪽 칼럼용 minipage (Traj. Error 2개 + VBench 4개)
    \begin{minipage}{0.51\textwidth}
        \centering
        \caption{
        \textbf{Stress test on conditioning trajectories.} \emph{Downsampled}: uniformly subsampled to \(N\) waypoints. The first and last waypoints are preserved. \emph{Perturbed}: Gaussian noise with standard deviation \(\sigma\) is added to each waypoint in the normalized \([-0.9, 0.9]\) coordinate space.
        } 
        \label{tab:abl-stress_test}
        \resizebox{\linewidth}{!}{%
        \begin{tabular}{l cc cccc}
            \toprule
            \multirow{2}{*}{\textbf{Trajectory Type}} 
            & \multicolumn{2}{c}{\textbf{Traj. Error} $\downarrow$}
            & \multicolumn{4}{c}{\textbf{VBench} $\uparrow$} \\
            \cmidrule(lr){2-3} \cmidrule(lr){4-7}
            & ADE\(_{3\text{D}}\) & FDE\(_{3\text{D}}\) & OC & AQ & MS & DD \\
            \midrule
            Original & \textbf{0.089} & \textbf{0.071} & \textbf{0.175} & \textbf{0.544} & 0.984 & \textbf{0.743} \\
            \midrule
            Downsampled \small{(\(N=8\))} & 0.093 & 0.080 & 0.174 & \textbf{0.544} & 0.985 & 0.729 \\
            Downsampled \small{(\(N=4\))} & 0.121 & 0.090 & 0.174 & 0.543 & \textbf{0.986} & 0.700 \\
            \midrule
            Perturbed \small{(\(\sigma=0.05\))} & 0.093 & 0.075 & \textbf{0.175} & \textbf{0.544} & 0.984 & \textbf{0.743} \\
            Perturbed \small{(\(\sigma=0.1\))} & 0.100 & 0.085 & 0.174 & 0.542 & 0.984 & \textbf{0.743} \\
            \bottomrule
            % Perturbed \small{(\(\sigma=0.2\))} & 0.135 & 0.131 & \textbf{0.175} & 0.536 & 0.984 & \textbf{0.786} \\
        \end{tabular}
        }
    \end{minipage}
\end{table*}
\subsection{Ablations}
\paragraph{Trajectory embedding method.}
To validate our shape-grounded trajectory embedding method, we compare it against two variants: (i) \emph{per-trajectory embedding}, which directly tokenizes each input trajectory and produces a variable number of condition tokens; and (ii) \emph{per-trajectory embedding with null padding}, which pads the null tokens to a fixed size to remove the effect of token sparsity. As shown in the top section of Tab.~\ref{tab:ablation_embedding}, our shape-grounded trajectory embedding achieves the lowest trajectory alignment error. It also obtains slightly higher VBench scores in aesthetic quality (AQ) and overall consistency (OC). 
The per-trajectory variant without padding obtains a higher dynamic degree (DD); however, this comes alongside worse trajectory alignment error, and the qualitative comparisons in Fig.~\ref{fig:traj-enc-abl-qual} show that the additional motion often appears in regions not specified by the input trajectories. This suggests that, when a spatially inconsistent and unevenly distributed trajectory set is naively injected, the model struggles to provide localized control over the intended regions. While this phenomenon is rarely observed in standard video generation with similar formulation, it emerges as a critical challenge in dynamic 3D generation. We suggest that this is because trajectory conditioning in video generation operates within a fixed 2D pixel space, where the background inherently provides dense, global guidance. Specifically, the background explicitly indicates unconditioned regions, and all trajectories are naturally spatially aligned within this grid. In our 3D setting, however, the spatial domain is highly shape-dependent, and the sparse input trajectories alone cannot provide sufficient global coverage across the entire surface.

% We attribute this in part to a property of the dynamic 3D generation task. Unlike video generation, where the trajectory condition operates in a fixed pixel space that is synchronized with the generation target and inherently covers the full output region through the background, our setting has neither of these properties: the target space varies with the input shape, and the partially given trajectories alone do not span the entire shape.
\paragraph{Effect of the number of anchor points.}
We further conduct an ablation study on the number of anchor points to examine its effect on performance. 
We report the results of varying \(N_{\text{anchor}}\) from 32 to 256 in the second section of Tab.~\ref{tab:ablation_embedding}. 
The performance remains stable across this range, indicating that our shape-grounded trajectory embedding is robust to the choice of \(N_{\text{anchor}}\). We set \(N_{\text{anchor}}\!=64\) in all experiments as it achieves strong trajectory alignment with comparable VBench scores.

% \paragraph{Varying the number of condition trajectory.}
% \lipsum[2]

\paragraph{Trajectory quality stress test.}
In real-world settings, user-provided trajectories may be noisy or coarse compared to the ground-truth trajectories used during training. 
To evaluate the robustness of our model against such imperfect inputs, we test two perturbations: (i) \emph{downsampled} trajectories, obtained by uniformly sampling \(N\) waypoints (\(N < 16\)) while preserving the first and last waypoints; and (ii) \emph{perturbed} trajectories, obtained by adding Gaussian noise with a standard deviation of \(\sigma\) to each waypoint (within a \([-0.9, 0.9]\) normalized coordinate space). In both cases, trajectory errors are computed against the original ground-truth trajectories. As reported in Tab.~\ref{tab:abl-stress_test}, while trajectory errors (ADE, FDE) inevitably increase, this degradation is marginal, and the video-based VBench metrics remain largely preserved. This indicates that the model robustly synthesizes the intended motion even when conditioned on coarse or moderately noisy trajectories.

\section{Conclusion}
% We presented \ours, a feed-forward framework for controllable dynamic 3D shape generation conditioned on 3D trajectories and a text prompt. 
% The key component is a shape-grounded trajectory embedding that maps an arbitrarily configured trajectory set into a dense, geometrically aware token set covering the entire shape. Across diverse evaluation, our framework produces motion that more faithfully reflects the given prompts than baselines, while supporting various applications and user-control signals such as sparse-trajectory dragging, motion editing, and motion transfer.

We presented \ours, a feed-forward framework for controllable dynamic 3D shape generation conditioned on 3D trajectories and a text prompt. The key component is a shape-grounded trajectory embedding that maps an arbitrarily configured trajectory set into a predefined shape-aware token set covering the entire shape, which enables stable conditioning under any input configuration. Across both quantitative and qualitative evaluations, our framework produces motion that more faithfully reflects the given prompts than the text-conditioned and video-based cascaded baselines, while supporting various motion controls and applications. We believe this approach makes a practical step toward enabling intuitive and expressive control in dynamic 3D shape generation.

\newpage
{
    \small
    \bibliographystyle{plainnat}
    \bibliography{reference}
}

\newpage
\appendix

\section*{Appendix}

% \subsection{Additional implementation details}
% \paragraph{Architecture details.}
% \textcolor{red}{shared attention weight}
\section{Implementation details}
\paragraph{Training details.}
Inspired by previous work on trajectory-guided generation~\cite{Zhang_2025_CVPR, yin2023dragnuwa, wang2024motionctrl}, we adopt a two-stage training strategy. In the first stage, trajectories are sampled at all \(N_{\text{anchor}}\) anchor points of the shape-grounded trajectory embedding, so that the model first learns to exploit the trajectory condition in a stable manner. In the second stage, the number of sampled trajectories is drawn uniformly at random between \(n_{\text{min}}\) and \(n_{\text{max}}\), where \(n_{\text{max}}\) is progressively annealed from \(N_{\text{anchor}}\) down to \(n_{\text{min}}\) over the course of training. To encourage the model to learn meaningful trajectory guidance, a fixed proportion \(\rho\) of the conditioning trajectories at each step is sampled exclusively from vertices with high motion magnitude, while the remaining trajectories are sampled uniformly at random across all vertices. Specifically, we draw these high-motion trajectories from the top \(30\%\) of vertices ranked by motion magnitude and use \(\rho = 0.7\).
% \paragraph{Architecture details.}
% We build on the pretrained B-spline-based VAE of~\cite{wang2025bimotion} and reuse the same VAE to embed the trajectory conditions for efficient training. The weights of the VAE encoder, including its two streams \(\mathcal{E}_{\text{shape}}\) and \(\mathcal{E}_{\text{traj}}\), are not updated during trajectory conditioning. We follow the original implementation for all transformations involved in the B-spline-based representation. Specifically, the cross-attention weight sharing in the original implementation is retained when computing the latent embeddings \(\mathbf{z}_{\text{shape}}\) and \(\mathbf{z}_{\text{traj}}\), but is omitted when encoding the trajectory condition to reduce the influence of the spatial configuration of the conditioning trajectories and the dilution introduced by the null embeddings. For brevity, we omit the cross-attention weight sharing in our notation. AdaLN~\cite{peebles2023scalable} conditioned on diffusion timestep is applied independently to the noisy latent \(\hat{\mathbf{z}}_{\text{traj}}\) and each conditioning token.

\paragraph{Architecture details.}
We build on the pretrained B-spline-based VAE of~\cite{wang2025bimotion} and reuse the same VAE to embed the trajectory conditions for efficient training. The weights of the VAE encoder, including its two streams \(\mathcal{E}_{\text{shape}}\) and \(\mathcal{E}_{\text{traj}}\), are not updated during trajectory conditioning. We follow the original implementation for all transformations involved in the B-spline-based representation. Specifically, the cross-attention weight sharing in the original implementation is retained when computing the latent embeddings \(\mathbf{z}_{\text{shape}}\) and \(\mathbf{z}_{\text{traj}}\), but is omitted when encoding the trajectory condition to reduce the influence of the spatial configuration of the conditioning trajectories and the dilution introduced by the null embeddings. With this cross-attention sharing disabled, the only remaining cross-trajectory interaction is removed, so the trajectory stream processes the conditioning trajectories in parallel without any coupling between them; encoding them jointly is therefore equivalent to encoding each one independently, which makes the per-trajectory embedding \(\mathcal{E}_{\text{traj}}(\tau_i)\) in Eq.~\ref{equ:shape-grounded_trajectory_embedding} well-defined regardless of how many conditioning trajectories are provided. The shape stream, in turn, encodes the input shape as in the original VAE, independently of the conditioning trajectories and anchors; the per-anchor shape feature \(\mathcal{E}_{\text{shape}}(s_i, n_i)\) denotes the resulting shape feature read out at the anchor location \((s_i, n_i)\). For brevity, we omit the cross-attention weight sharing in our notation. AdaLN~\cite{peebles2023scalable} conditioned on diffusion timestep is applied independently to the noisy latent \(\hat{\mathbf{z}}_{\text{traj}}\) and each conditioning token.

\paragraph{Anchored farthest point sampling.}
Our shape-grounded trajectory embedding requires a fixed set of \(N_{\text{anchor}}\) anchor points that (i) include the starting positions of all user-provided trajectories so that each trajectory is grounded at its intended location on the shape, and (ii) cover the rest of the shape to provide geometric context for unconditioned regions. We achieve both with an anchored variant of farthest point sampling. Given a trajectory set \(\mathcal{T} = \{{\tau_k}\}_{k=1}^{|\mathcal{T}|}\), we first take the starting waypoints \(\{{\mathbf{p}_{k,0}}\}_{k=1}^{|\mathcal{T}|}\) as the trajectory anchors and include them as the first \(|\mathcal{T}|\) entries of the anchor set. The remaining \(N_{\text{anchor}} - |\mathcal{T}|\) anchors are then selected from the mesh vertices \(\mathbf{V}_0\) by farthest point sampling seeded with the trajectory anchors, so that newly added anchors are progressively chosen to maximize the minimum distance to the existing anchor set. During training, each trajectory anchor coincides with a mesh vertex; at inference, when a user-provided trajectory does not start exactly on a vertex, its starting position is snapped to the nearest point on the mesh surface before serving as an anchor. This procedure guarantees that the trajectory anchors are preserved exactly while the remaining anchors evenly cover the shape.

% \paragraph{Training details.}

% \textcolor{red}{iteration of augmentation}

% \textcolor{red}{large motion sampling}

\section{Data augmentation} \label{appendix:data-aug}
To improve robustness and encourage the model to actively utilize trajectory information, we apply two categories of data augmentations. All augmentations described below are applied only when trajectories are provided as conditions, since they perturb the data distribution that text alone could otherwise capture and may degrade text-conditioned generation. First, we apply global geometric and temporal augmentations. We inject random noise into the input shape (positions and normals) as well as the target and conditioning trajectories, alongside random horizontal rotation and flipping. Additionally, we apply temporal cropping that retains at least 50\% of the full sequence length to preserve the original motion semantics. These transformations not only enhance overall robustness but also encourage the model to resolve spatial ambiguities—which cannot be inferred from text alone—by relying on the trajectory guidance. Second, to ensure the model is robust to the coarse and noisy trajectories typically provided by users, we apply targeted augmentations exclusively to the conditioning trajectories. This involves injecting Gaussian noise, followed by two smoothing techniques: a simplification step that subsamples a random subset of intermediate waypoints while retaining the endpoints, and 1D Gaussian-kernel smoothing applied along the temporal axis with a randomly chosen kernel width, which suppresses high-frequency jitter and yields smoother trajectories closer to those a user would draw.
% To improve robustness and encourage the model to actively utilize trajectory information, we apply two categories of data augmentations. First, we apply global geometric and temporal augmentations. We inject random noise into the input shape (positions and normals) as well as the target and conditioning trajectories, alongside random horizontal rotation and flipping. Additionally, we apply temporal cropping that retains at least 70\% of the full sequence length to preserve the original motion semantics. These transformations not only enhance overall robustness but also encourage the model to resolve spatial ambiguities—which cannot be inferred from text alone—by relying on the trajectory guidance. Second, to ensure the model is robust to the coarse and noisy trajectories typically provided by users, we apply targeted augmentations exclusively to the conditioning trajectories. This involves injecting Gaussian noise, followed by two smoothing techniques: a simplification step that subsamples intermediate waypoints while retaining the endpoints, and 1D Gaussian-kernel smoothing.
\begin{figure}[t!]
    % \centering
    % 가로 길이는 \textwidth, 세로는 그 2/3인 0.66\textwidth로 설정
    \includegraphics[width=\textwidth]{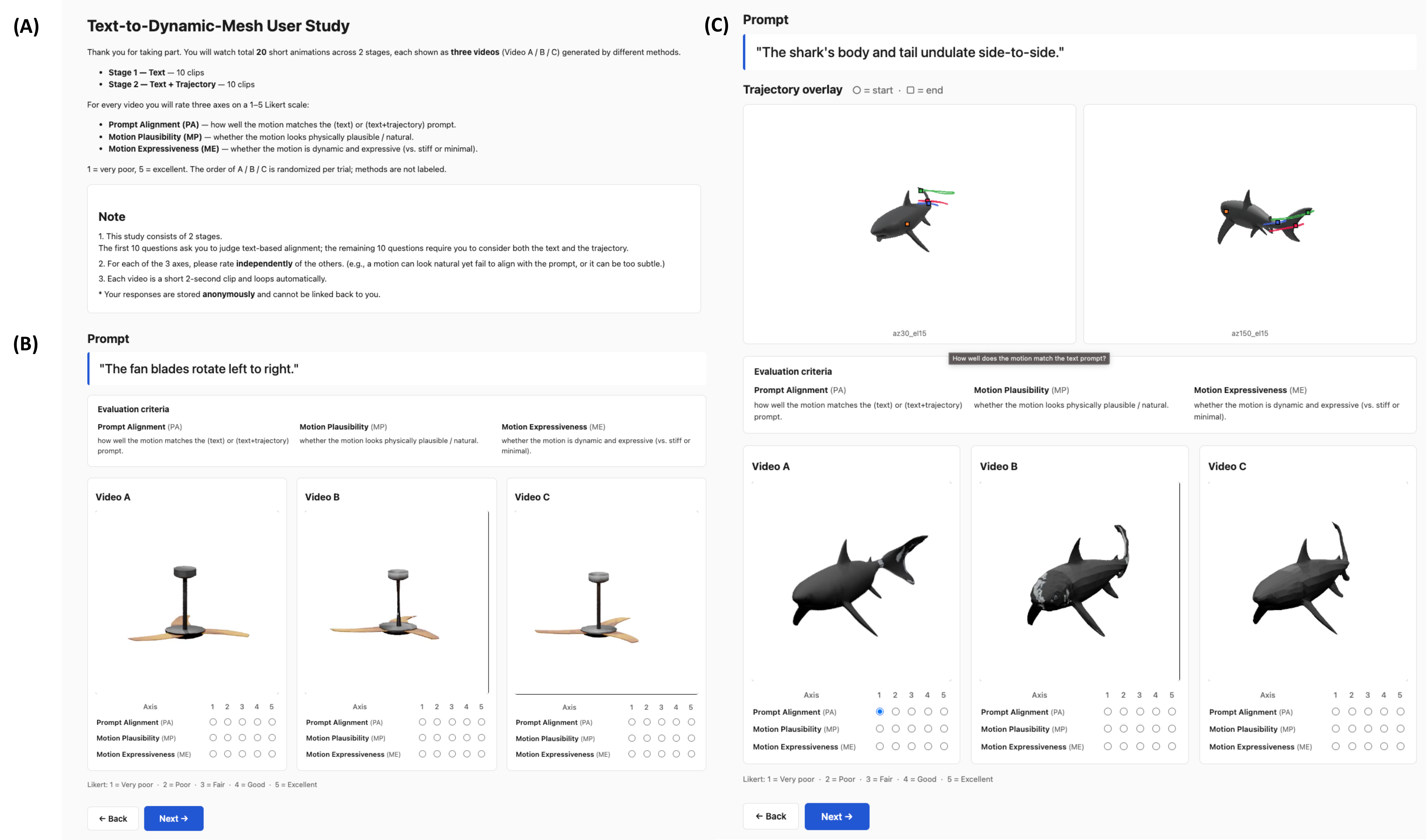}
    \caption{
    \textbf{User study interface.} \textbf{(A)} Initial instruction page describing the rating task and evaluation criteria. \textbf{(B)} Rating page for text-conditioned baselines, where participants rate three videos generated by different methods given a text prompt. \textbf{(C)} Rating page for cascaded video-based baselines, where participants additionally see the input trajectory overlay alongside the text prompt. Videos from compared methods are presented side-by-side in a randomly shuffled order.
    }
    \label{fig:userstudy-screenshot}
\end{figure}

\begin{table*}[t]
    \centering
    \caption{\textbf{User study results.} Each cell shows the mean \(\pm\) standard deviation over all participants and examples. \emph{Left}: text-based baselines~\cite{Wu_2025_ICCV, wang2025bimotion}. \emph{Right}: video-based baselines~\cite{Zhang_2025_CVPR, sabathier2026actionmesh, chen2026motion}.}
    \label{tab:user_study_results}

    \begin{minipage}{0.48\textwidth}
        \centering
        \resizebox{\linewidth}{!}{%
        \begin{tabular}{lccc}
            \toprule
            \textbf{Methods} & \textbf{PA\(\uparrow\)} & \textbf{MP\(\uparrow\)} & \textbf{ME\(\uparrow\)} \\
            \midrule
            AnimateAnyMesh~\cite{Wu_2025_ICCV} & 2.170\(\pm\)1.260 & 2.667\(\pm\)1.452 & 2.560\(\pm\)1.243 \\
            BiMotion~\cite{wang2025bimotion}       & 2.657\(\pm\)1.525 & 3.040\(\pm\)1.536 & 2.813\(\pm\)1.499 \\
            \rowcolor{cyan!10}
            \textbf{Ours}  & \textbf{4.507}\(\pm\)\textbf{0.782} & \textbf{4.067}\(\pm\)\textbf{1.055} & \textbf{4.340}\(\pm\)\textbf{0.880} \\
            \bottomrule
        \end{tabular}%
        }
    \end{minipage}
    \hfill
    \begin{minipage}{0.48\textwidth}
        \centering
        \resizebox{\linewidth}{!}{%
        \begin{tabular}{lccc}
            \toprule
            \textbf{Methods} & \textbf{PA\(\uparrow\)} & \textbf{MP\(\uparrow\)} & \textbf{ME\(\uparrow\)} \\
            \midrule
            ActionMesh~\cite{sabathier2026actionmesh}   & 1.780\(\pm\)1.004 & 1.687\(\pm\)1.042 & 2.040\(\pm\)1.053 \\
            Motion 3-to-4~\cite{chen2026motion} & 1.940\(\pm\)1.071 & 1.800\(\pm\)1.025 & 1.903\(\pm\)0.944 \\
            \rowcolor{cyan!10}
            \textbf{Ours} & \textbf{4.577}\(\pm\)\textbf{0.657} & \textbf{4.177}\(\pm\)\textbf{0.921} & \textbf{4.437}\(\pm\)\textbf{0.775} \\
            \bottomrule
        \end{tabular}%
        }
    \end{minipage}
\end{table*}

\section{Evaluation details} \label{appendix:eval-detail}
\paragrapht{Evaluation metrics.} \label{appendix:metric}
We represent each generated motion as a sequence of \(T\) deformed meshes.
Let \(\mathbf{V}_{t} \in \mathbb{R}^{N \times 3}\) denote the vertex positions of the generated mesh at frame \(t \in \{1,\dots,T\}\), where \(N\) is the number of vertices and \((\mathbf{V}_{t})_{i}\) is the 3D position of vertex \(i\).

For evaluation, we are given a mesh with motion together with \(K\) target trajectories used as conditioning. Each trajectory has its source point on the mesh surface, where the source point is included in mesh vertices, i.e., \(\mathbf{p}_{k,0} \in \mathbf{V}_{0}\). We denote the corresponding vertex index by \(i_k\), so that \((\mathbf{V}_{0})_{i_k} = \mathbf{p}_{k,0}\). The generated vertex \((\mathbf{V}_{t})_{i_k}\) is then compared with the target waypoint \(\mathbf{p}_{k,t}\) at each frame.

\textit{Average Motion Distance (AMD)} measures the average inter-frame displacement over all vertices:
\begin{equation}
    \text{AMD} =
    \frac{1}{T-1}\sum_{t=1}^{T-1}
    \frac{1}{N}\sum_{i=1}^{N}
    \left\lVert
    (\mathbf{V}_{t+1})_{i} - (\mathbf{V}_{t})_{i}
    \right\rVert_{2}.
\end{equation}
AMD measures the overall amount of generated motion, not trajectory-tracking accuracy. It is computed in 3D world coordinates and reported in the same units as the mesh.

\textit{Average Displacement Error (ADE)} and \textit{Final Displacement Error (FDE)} measure trajectory-tracking accuracy for the \(K\) controlled vertices. For the \(k\)-th trajectory, we define
\begin{equation}
    \text{ADE}_{k} =
    \frac{1}{T}\sum_{t=1}^{T}
    \left\lVert
    (\mathbf{V}_{t})_{i_k} - \mathbf{p}_{k,t}
    \right\rVert_{2},
    \qquad
    \text{FDE}_{k} =
    \left\lVert
    (\mathbf{V}_{T})_{i_k} - \mathbf{p}_{k,T}
    \right\rVert_{2}.
\end{equation}
The final ADE and FDE are obtained by averaging over all input trajectories:
\begin{equation}
    \text{ADE} = \frac{1}{K}\sum_{k=1}^{K}\text{ADE}_{k},
    \qquad
    \text{FDE} = \frac{1}{K}\sum_{k=1}^{K}\text{FDE}_{k}.
\end{equation}

For the ADE\(_\text{3D}\) and FDE\(_\text{3D}\), the above distances are computed directly in world coordinates. For the 2D variant, both the generated vertex and the target waypoint are projected to the image plane using the evaluation camera projection \(\pi:\mathbb{R}^{3}\to\mathbb{R}^{2}\):
\begin{equation}
    \text{ADE}_{\text{2D},k} =
    \frac{1}{T}\sum_{t=1}^{T}
    \left\lVert
    \pi\!\left((\mathbf{V}_{t})_{i_k}\right)
    -
    \pi\!\left(\mathbf{p}_{k,t}\right)
    \right\rVert_{2},
    \qquad
    \text{FDE}_{\text{2D},k} =
    \left\lVert
    \pi\!\left((\mathbf{V}_{T})_{i_k}\right)
    -
    \pi\!\left(\mathbf{p}_{k,T}\right)
    \right\rVert_{2}.
\end{equation}
The 2D errors are reported in pixels and are averaged over the fixed evaluation viewpoints.

\paragraph{Condition trajectory selection.} 
For quantitative evaluation, we obtain the input trajectories for our model automatically, without human intervention, by sampling from the per-vertex ground-truth trajectories using a two-stage predefined rule-based procedure. This process is performed to avoid bias introduced by subjective trajectory selection. In the first stage, we score each vertex by its cumulative path length over the sequence and draw \(N_{\text{cand}} = 64\) candidates from the top-30\% high-motion vertices, with sampling probabilities biased toward higher-motion vertices by a power law with exponent \(\beta = 2\). In the second stage, the candidates are clustered using K-Means on a feature composed of each vertex's source position and its per-frame displacements. The number of clusters \(K_{c} \in \{3, 4, 5\}\) is selected by the silhouette score, and the trajectory closest to each cluster centroid is retained as the representative. The resulting set \(\mathcal{R}\), with \(|\mathcal{R}| = K_{c} \le 5\), is used as the trajectory input for quantitative evaluation. For qualitative figures, we use the same automatic procedure unless otherwise specified. When specified trajectories are used, we explicitly visualize the conditioning trajectories in the figure. The average number of conditioning trajectories is \(4.01\) in quantitative evaluation, and no hyperparameter search was performed to optimize the evaluation results.

\paragraph{User study protocol.}
Since subtle aspects of motion are difficult to evaluate with automatic pipelines, we conduct an anonymous user study to obtain human perceptual judgments. The study involves 30 participants, each rating the generated animations along three axes: \emph{Prompt Alignment} (PA), \emph{Motion Plausibility} (MP), and \emph{Motion Expressiveness} (ME). Ratings are reported on a 5-point Likert scale, where 1 denotes \emph{very poor} and 5 denotes \emph{excellent}.
We evaluate against two groups of baselines: text-conditioned baselines~\cite{Wu_2025_ICCV, wang2025bimotion} that take only text as input, and cascaded video-based baselines~\cite{Zhang_2025_CVPR, sabathier2026actionmesh, chen2026motion} that take both text and trajectories. For each group, we independently sample 10 objects at random from the evaluation set. For the text-based comparison, prompt alignment (PA) is assessed against the text prompt. For the video-based baselines, PA is assessed against both the trajectory condition and the text prompt, since these models receive both conditions. In total, each participant answers \((10+10) \times 3 \times 3 = 180\) questions, covering 20 objects, three methods, and three rating axes. As shown in Fig.~\ref{fig:userstudy-screenshot}, participants are first presented with the rating instructions, after which they rate each object using a video and the corresponding prompt. For each object, the videos from all compared methods are presented side-by-side in a randomly shuffled order so that participants are blind to the source of each video, and each video is automatically looped. A consolidated summary of the user study results, including standard deviations, is reported in Tab.~\ref{tab:user_study_results}.
% No IRB approval or formal exemption was obtained.
% The user study was conducted anonymously. Participants were informed of the study procedure before participation, participated voluntarily, and could stop at any time. We did not collect personally identifiable information or sensitive personal data. The study involved minimal risk, as participants only viewed generated animations and provided Likert-scale ratings. 
% Each participant received a compensation after completing the study. The study took approximately 15 minutes per participant, and the compensation was set to be at least comparable to the applicable local minimum wage on an hourly basis.

\begin{figure}[t!]
    \centering
    % 가로 길이는 \textwidth, 세로는 그 2/3인 0.66\textwidth로 설정
    \includegraphics[width=\textwidth]{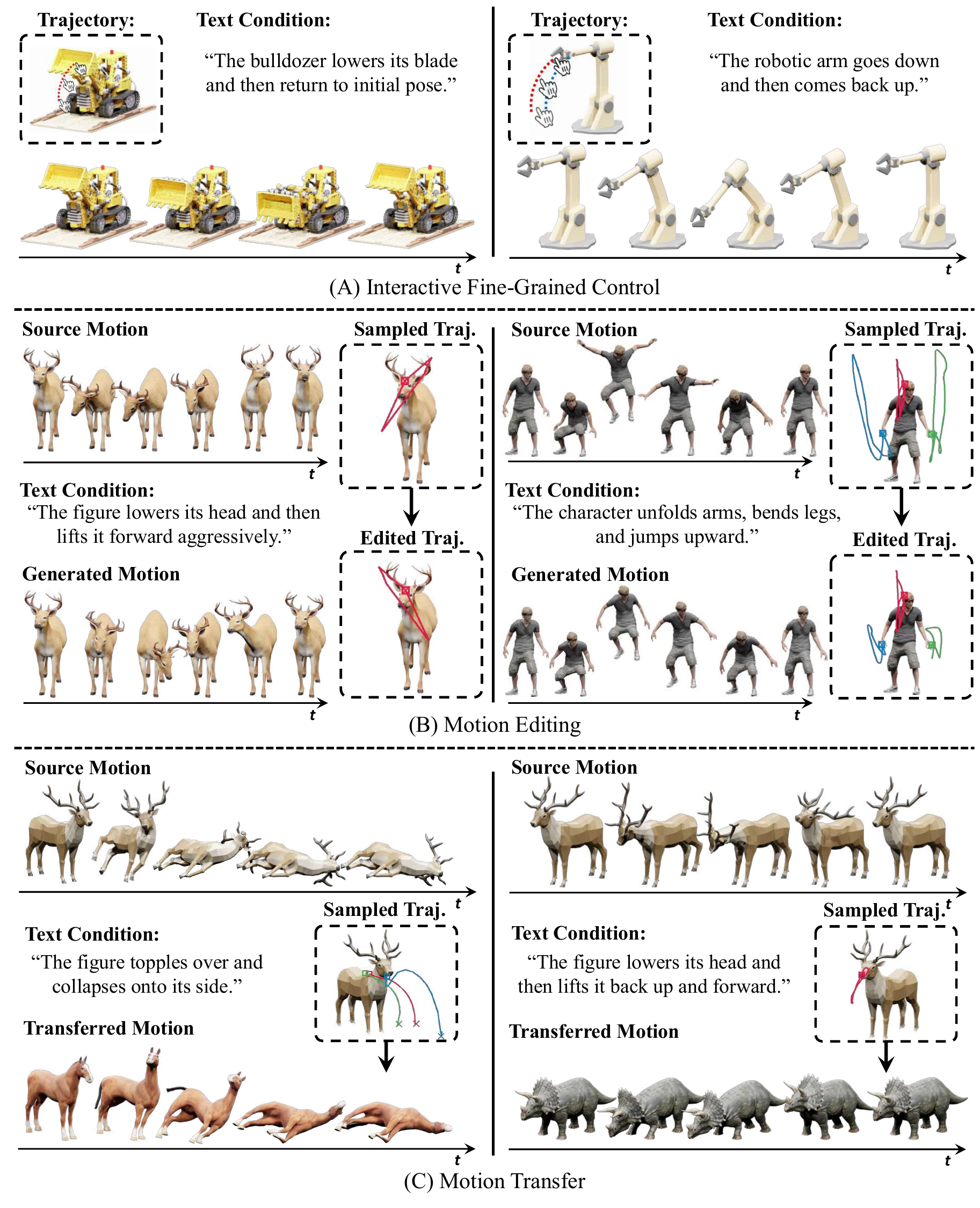}
    \caption{
    \textbf{Application showcase.} Our framework supports a diverse range of motion controls through joint trajectory and text conditioning. \textit{Top}: fine-grained motion control with user-specified trajectories. \textit{Middle}: motion editing by modifying a few waypoints of representative trajectories. \textit{Bottom}: motion transfer from a source motion to a target mesh via sampled representative trajectories. Our model uses only the visualized trajectories as input.
    }
    \label{fig:applications}\vspace{-10pt}
\end{figure}

\paragrapht{Video generation procedure.}
To evaluate the cascaded video-based baselines which combines trajectory-conditioned video generation model with video-to-dynamic mesh generation model, we first generate trajectory-conditioned videos using Tora~\cite{Zhang_2025_CVPR}. We choose Tora as the video-generation baseline because it is a publicly available trajectory-conditioned video generator that supports arbitrary point-level 2D trajectory control, rather than relying on object-level controls. Since Tora operates in the image plane, we project the input 3D trajectories into 2D pixel coordinates using the same camera viewpoint as the rendered videos used for VBench~\cite{huang2024vbench}. To avoid degraded video quality caused by occluded trajectories, we discard any trajectory whose source vertex is occluded in the first frame; occlusion is determined by comparing the projected depth of each vertex with the depth value at the corresponding pixel. The resulting videos, together with the ground-truth source mesh, are then passed to the video-to-dynamic-3D mesh generation pipelines~\cite{sabathier2026actionmesh, chen2026motion}, which produce the final dynamic mesh sequence used for evaluation. All baselines are run with their publicly available models and default configurations.
\section{Applications}
Trajectories provide a flexible medium for expressing various forms of motion control. By accepting trajectory sets of arbitrary configuration, our model exploits this flexibility to enable diverse applications, including interactive fine-grained control, motion editing, and motion transfer.

\paragraph{Interactive fine-grained control.} 
It is challenging to describe the precise motion which a user imagines, and as shown in Fig.~\ref{fig:qual-comparison-w-text}, models conditioned solely on text often struggle to faithfully reflect detailed prompts. In contrast, our model accepts explicit spatial trajectories, allowing users to specify motion attributes such as direction and magnitude. As shown in the first row of Fig.~\ref{fig:applications}, by reflecting user-provided trajectories together with the global semantics of the text prompt, our model enables fine-grained motion generation in an interactive manner, specifying the direction and magnitude.

\paragraph{Motion editing.}
Refining a previously generated or pre-existing motion is a time-consuming and laborious task: without rigging, it requires directly adjusting the movements of many individual vertices, and even with rigging, multiple skeletal motions must be coherently edited manually. Since our model can generate motions consistent with a sparse set of input trajectories, editing reduces to modifying a few waypoints of a small number of representative trajectories in the region of interest. Specifically, given a set of representative trajectories \(\mathcal{T}_{\text{rep}} = \{\tau_{k}\}_{k=1}^{|\mathcal{T}_{\text{rep}}|}\) sampled from the source motion, the user edits a subset of waypoints indexed by
\(\mathcal{I} \subseteq \{1,\dots,|\mathcal{T}_{\text{rep}}|\} \times \{1,\dots,T\}\).
For each \((k,t)\in\mathcal{I}\), the original waypoint \(\mathbf{p}_{k,t}\) is replaced with an edited waypoint \(\mathbf{p}_{k,t}^{\star}\), while all other waypoints are kept unchanged. This yields an edited trajectory set \(\mathcal{T}_{\text{rep}}^{\star}\). We then pass \(\mathcal{T}_{\text{rep}}^{\star}\), the text prompt, and a target frame of the source motion as the source mesh to our model, which generates the edited motion. We show examples in the second row of Fig.~\ref{fig:applications}.

\paragraph{Motion transfer.} 
Transferring a source motion to another mesh in a spatially consistent manner is difficult to achieve with text alone, as text cannot impose precise spatial constraints, and is also unstable with video-based approaches due to topology mismatch. In contrast, 3D trajectories provide a topology-agnostic motion representation, since they specify point-wise paths in space rather than depending on the mesh connectivity or the reference topology. By sampling a few representative trajectories from a source object and attaching them to the corresponding vertices of a target mesh, our model performs motion transfer without additional supervision. Specifically, given a source motion defined on a mesh \(\mathcal{M}_{\text{src}} = (\mathbf{V}_{\text{src}}, \mathbf{F}_{\text{src}})\) with per-vertex trajectories \(\mathcal{T}_{\text{src}} = \{\tau_{k}\}_{k=1}^{|\mathbf{V}_{\text{src}}|}\), we sample a sparse subset \(\mathbf{V}_{\text{src}}^{\star} \subset \mathbf{V}_{\text{src}}\) of representative vertices and their associated trajectories \(\mathcal{T}_{\text{src}}^{\star} \subset \mathcal{T}_{\text{src}}\). We attach \(\mathcal{T}_{\text{src}}^{\star}\) to the corresponding vertices \(\mathbf{V}_{\text{tgt}}^{\star}\) of the target mesh \(\mathcal{M}_{\text{tgt}} = (\mathbf{V}_{\text{tgt}}, \mathbf{F}_{\text{tgt}})\). Here, the correspondence between \(\mathbf{V}_{\text{src}}^{\star}\) and \(\mathbf{V}_{\text{tgt}}^{\star}\) is specified manually by selecting semantically matching points (e.g., corresponding limbs or other shared parts) on the two meshes; since the representative subset is sparse—only three points in our examples—this requires minimal user effort. We then feed the text prompt, the attached trajectories, and \(\mathcal{M}_{\text{tgt}}\) to our model to transfer the source motion onto the target mesh. We showcase examples in the last row of Fig.~\ref{fig:applications}, where only three representative trajectories are sampled from the source motion.
\section{Social impact and limitations}

\paragraph{Social impact.}
This work aims to make dynamic 3D content creation more accessible by enabling users to control mesh motion through sparse 3D trajectories and natural language. Such controllability can reduce the amount of manual labor required for animation, particularly for users without expertise in rigging, keyframing, or skeletal motion editing, thereby supporting more efficient workflows in digital content creation, AR/VR, gaming, filmmaking, education, and simulation. At the same time, controllable dynamic 3D generation may introduce risks. Generated animated assets could be misused to create deceptive, misleading, or harmful visual content, especially when combined with realistic rendering. The ability to edit or transfer motion may also raise concerns about unauthorized reuse of copyrighted assets, character designs, or motion data. In addition, if training data contain biased or inappropriate content, generated motions may inherit or amplify such biases. These risks highlight the importance of responsible data curation, compliance with dataset licenses, clear usage guidelines, and provenance mechanisms such as watermarking or metadata for generated assets. We encourage release practices that include documentation of training data sources, intended use cases, and known limitations.

\paragraph{Limitations.}
Although T2Mo offers a high degree of controllability via 3D trajectories and text, our formulation predicts per-vertex displacements on a fixed source mesh and therefore assumes a constant topology across frames. Consequently, motions involving topological changes, such as splitting, merging, or the emergence of new components, lie outside the scope of our model. A future direction is to integrate our shape-grounded trajectory embedding with per-frame generation methods, enabling trajectory-controllable dynamic 3D generation under varying topology, thereby extending our framework beyond motion synthesis to broader dynamic scene generation where the underlying geometry itself evolves over time.

% \input{Suppl/MoreResults}

% \input{Suppl/TrajAugPseudoCode}

% Technical appendices with additional results, figures, graphs, and proofs may be submitted with the paper submission before the full submission deadline (see above). You can upload a ZIP file for videos or code, but do not upload a separate PDF file for the appendix. There is no page limit for the technical appendices. 

% Note: Think of the appendix as ``optional reading'' for reviewers. The paper must be able to stand alone without the appendix; for example, adding critical experiments that support the main claims to an appendix is inappropriate. 

%%%%%%%%%%%%%%%%%%%%%%%%%%%%%%%%%%%%%%%%%%%%%%%%%%%%%%%%%%%%

% \newpage
% \input{checklist.tex}

\end{document}